\documentclass{article}

\usepackage{arxiv}

\usepackage[utf8]{inputenc} 
\usepackage[T1]{fontenc}  
\usepackage{hyperref}    
\usepackage{url}      
\usepackage{booktabs}    
\usepackage{amsfonts}    
\usepackage{nicefrac}    
\usepackage{microtype}   
\usepackage{lipsum}
\usepackage{graphicx}
\usepackage{times}
\usepackage{hyperref}
\usepackage{url}
\usepackage{xcolor} 
\definecolor{LightGray}{gray}{0.9}
\usepackage{minted}
\usepackage{graphicx}
\usepackage{subcaption}
\usepackage{bbm}
\usepackage{mathtools}
\usepackage[dvipsnames]{xcolor}
\usepackage{booktabs}
\usepackage{adjustbox}
\usepackage{multirow}
\usepackage{makecell}
\usepackage{algorithm2e}
\usepackage{natbib}

\title{On the Limits of Tabular Hardness Metrics for Deep RL: A Study with the Pharos Benchmark}

\input{math_commands.tex}

\author{
 Michelangelo Conserva \quad Remo Sasso \quad Paulo Rauber \\
 School of Electronic Engineering and Computer Science\\
 Queen Mary University of London, United Kingdom \\
 \texttt{\{m.conserva, r.sasso, p.rauber\}@qmul.ac.uk}
}

\begin{document}
\maketitle

\begin{abstract}
Principled evaluation is critical for progress in deep reinforcement learning (RL), yet it lags behind the theory-driven benchmarks of tabular RL. While tabular settings benefit from well-understood hardness measures like MDP diameter and suboptimality gaps, deep RL benchmarks are often chosen based on intuition and popularity. This raises a critical question: can tabular hardness metrics be adapted to guide non-tabular benchmarking? We investigate this question and reveal a fundamental gap. Our primary contribution is demonstrating that the difficulty of non-tabular environments is dominated by a factor that tabular metrics ignore: representation hardness. The same underlying MDP can pose vastly different challenges depending on whether the agent receives state vectors or pixel-based observations. To enable this analysis, we introduce \texttt{pharos}, a new open-source library for principled RL benchmarking that allows for systematic control over both environment structure and agent representations. Our extensive case study using \texttt{pharos} shows that while tabular metrics offer some insight, they are poor predictors of deep RL agent performance on their own. This work highlights the urgent need for new, representation-aware hardness measures and positions \texttt{pharos} as a key tool for developing them.
\end{abstract}

\section{Introduction}

Progress in reinforcement learning (RL) is fundamentally tied to the rigorous evaluation of agent performance. While deep RL agents have demonstrated remarkable success in complex simulated domains like board games and video games \citep{silver2018general, vinyals2019grandmaster}, their high sample complexity remains a significant barrier to deployment in real-world applications such as finance, healthcare, and robotics. In these domains, interactions with the environment are often expensive, time-consuming, or carry inherent risks, making the collection of vast amounts of data infeasible. Consequently, improving the sample efficiency of RL agents is critical.

Much of the research in RL is dedicated to this goal, either by developing more efficient exploration strategies \citep{osband2019deep} or by integrating prior knowledge into the learning process \citep{llm_survey}. However, the validation of these advances depends entirely on the quality of our evaluation methodologies. This is a non-trivial challenge, complicated by environmental stochasticity \citep{machado2018revisiting}, inherent variability in agent behavior \citep{jordan2020evaluating}, and the difficulty of defining universally meaningful performance metrics \citep{mendez2022composuite}. A rigorous evaluation process, built upon well-designed benchmarks, is therefore an essential prerequisite for measuring progress and ensuring that reported improvements are both significant and generalizable.

The core of this evaluation process lies in the design of benchmarks. Ideally, a benchmark should present a diverse range of challenges, be compelling enough to engage the research community, and provide clear insights into the strengths and weaknesses of different agents. In the tabular RL setting, significant strides have been made toward this ideal by leveraging a well-established theory of hardness \citep{conserva2022hardness}. This allows for the principled construction of benchmarks with environments selected for their theoretically characterized difficulty profiles.

In contrast, the non-tabular setting lacks a practically applicable theory of hardness. While theoretical constructs exist, they often rely on restrictive assumptions or are computationally intractable for the complex, high-dimensional environments that are of primary interest to the community. As a result, non-tabular benchmarking relies on expert judgment and community popularity, which, while useful, lack the rigor needed to systematically probe agent capabilities. This gap between the principled approach in tabular RL and the ad-hoc nature of non-tabular RL motivates our work.

This paper takes a first step toward the ambitious goal of designing principled benchmarks for non-tabular reinforcement learning. Our investigation proceeds in three parts. We first survey the landscape of hardness theory in Section~\ref{sec:hardness}, contrasting the mature theory of the tabular setting with the practical limitations of its non-tabular counterpart. This review highlights that existing measures fail to account for the critical challenge of representation learning. To address the practical needs for such an investigation, we then outline the desiderata for a principled benchmarking tool in Section~\ref{sec:bench} and introduce our solution: \texttt{pharos}. This new library is specifically designed to bridge the gap between tabular analysis and non-tabular complexity by supporting both scalable hardness computation and fine-grained control over environments and agent observations. With this tool in hand, we conduct an extensive case study in Section~\ref{sec:exp} using tabular Q-learning and DQN. The results reveal that agent performance is critically dependent on the input representation, often more so than on the underlying tabular difficulty. This leads to our central conclusion: while tabular measures offer a starting point, the path to principled non-tabular benchmarking requires new metrics designed to quantify the unique challenges of representation.

\section{Hardness in Reinforcement Learning} \label{sec:hardness}

Section~\ref{ref:preliminaries} formally introduces the reinforcement learning setting.
Section~\ref{sec:tabular} reviews the tabular hardness metrics that can be computed in practice.
Section~\ref{sec:non-tabular} highlights the key challenges in deriving a practically useful theory of hardness in the non-tabular setting.
Section~\ref{sec:open-problems} formalizes the open problems and suggests future research directions.

\subsection{Preliminaries} \label{ref:preliminaries}

We formulate an RL problem as a Markov decision process (MDP) $\mathcal{M} = (\mathcal{S}, \mathcal{A}, p, r, \gamma)$ composed of
a set of states $\mathcal{S}$;
a finite set of actions $\mathcal{A}$;
a transition model $p : \mathcal{S} \times \mathcal{A} \times \mathcal{S} \to [0, 1]$ such that $\sum_{s'} p(s, a, s') = 1$ for every $(s, a) \in \mathcal{S} \times \mathcal{A}$;
a reward function $r: \mathcal{S} \times \mathcal{A} \to \mathbb{R}$;
and a discount factor $\gamma \in (0, 1)$.
If for every $(s, a) \in \mathcal{S} \times \mathcal{A}$ there is a state $s' \in \mathcal{S}$ such that $p(s, a, s') = 1$ , we say that the transition model $p$ is deterministic.

The interaction between the agent and the environment starts at time step $t=0$ and ends at time $t=T$.
For any $t \geq 0$, the agent selects an action $a_t$ from its policy $\pi: \mathcal{S} \to \mathcal{A}$ and the environment draws reward $r_{t+1}$ and next state $s_{t+1}$.
For the episodic setting, we refer to the episode length as $H$.

The objective of a reinforcement learning agent is to find an optimal policy $\pi^*$ that maximizes the expected discounted return defined as $G^{\pi}_t = \mathbb{E}_{\pi} \left[ \sum_{k = 0}^\infty \gamma^{k} r(s_t, \pi(s_t)) \right]$.
The state-value function $V^{\pi}(s) = \mathbb{E}_{\pi} \left[ G_t | S_t = s \right]$ quantifies the expected return when starting in state $s$ and following policy $\pi$. 
Similarly, the action-value function $Q^{\pi}(s, a) = \mathbb{E}_{\pi} \left[ G_t | S_t = s, A_t = a \right]$ quantifies the expected return when starting in state $s$, taking action $a$, and then following policy $\pi$. The optimal state-value and action-value functions are denoted by $V^*$ and $Q^*$, respectively.

\subsection{Tabular} \label{sec:tabular}

In the tabular setting, the agent can maintain distinct value estimates for each state-action pair, facilitating systematic exploration. However, this also necessitates visiting a substantial portion of the state-action space, as the value of one state-action pair $(s, a)$ provides no direct information about the value of another $(s', a')$. For example, in a tree-structured episodic MDP, if $s$ and $s'$ are leaf nodes, their values are independent. While complete exploration is not strictly required due to bounded rewards and the possibility of pruning based on maximum return, the theoretical lower bound on regret (e.g., $\sqrt{HSAT}$ in the episodic case \citep{dann2015sample}) reflects the dependence on the number of states and actions.

\textbf{Hardness characterization.}
To study the hardness of this setting, MDP complexity has been broken down into two categories: \textbf{visitation complexity} and \textbf{estimation complexity} \citep{conserva2022hardness}.
Visitation complexity refers to the difficulty an agent faces in visiting all states, which is crucial for learning an effective policy.
Estimation complexity, on the other hand, deals with the challenges of estimating the optimal policy accurately based on the samples gathered, highlighting the discrepancy between the true optimal policy and the best policy derived from the given estimates of the transition and reward kernels.
These two categories are complementary; visitation complexity focuses on the effort required to explore the state space thoroughly, while estimation complexity concerns the precision needed in policy estimation given the gathered data.

\textbf{Diameter.}
The diameter of an MDP provides a measure of its visitation complexity, quantifying the difficulty of moving between states. Formally, the diameter D is defined as:
\begin{equation}
  D := \sup\limits_{s' \in \mathcal{S}} \inf_\pi T^\pi_{s \to s'},
\end{equation}
where represents the expected number of steps to reach state $s$ from $s'$ under policy $\pi$ \citep{jaksch2010near}.
While the diameter effectively captures the challenge of gathering samples from the environment, it fails to account for the reward structure and the resulting value functions. Consequently, it doesn't address the estimation complexity of the MDP.

\textbf{Suboptimality gaps.}
The sum of the reciprocals of the suboptimality gaps offers a measure of estimation complexity \citep{simchowitz2019non}. These gaps are defined as:
\begin{equation}
  \Delta(s,a) := V^*(s) - Q^*(s,a)
\end{equation}
Since $V^*(s) = \max_a Q^*(s,a)$ for all states $s$, the suboptimality gap $\Delta(s,a)$ quantifies the difference in expected return between taking the optimal action and taking action $a$ in state $s$. Intuitively, a larger gap $\Delta(s,a')$ for a suboptimal action $a'$ in state $s$ makes it easier to identify $a'$ as suboptimal. Unlike the diameter, this measure does not consider the difficulty of potentially hard-to-reach states, and, instead, it focuses on the complexity of accurately estimating the optimal policy.

\subsection{Non-tabular} \label{sec:non-tabular}

Contrary to the tabular setting, in non-tabular MDPs, the value function of state action pairs can be related even if state $s$ is not a predecessor of state $s'$ or vice versa.
For example, in the Atari game Freeway (where the agent needs to cross a road), if the agent learns that, to maximize the total score, it needs to avoid being hit by a car, then it can transfer this knowledge to states that are not related from a tabular point of view, e.g. cars from different lanes.
From a theoretical perspective, this relationship is formalized by assuming the functional form of the transition and reward functions.

\textbf{Feature dimension.}
A well-studied non-tabular MDP model is the Low-Rank MDP \citep{zanette2019tighter, agarwal2020flambe}, which assumes a linear functional form in the reward and transition dynamics.
Formally, there exists a known feature map $\phi : \cS \times \cA \to \bR^d$ and parameters $\theta^r \in \bR^d$, $\theta^P \in \bR^d$ such that
\begin{gather}
  r(s,a) = \phi(s, a)^T \theta^r \quad \text{and} \quad p(s, a, s') = \phi(s, a)^T \psi(s').
\end{gather}
We are also guaranteed that for every policy $\pi$ there exists a vector $\theta^\pi \in \bR^d$ such that,
\begin{gather}
  Q^\pi(s,a) = \phi(s,a)^T \theta^\pi.
\end{gather}
While this setting is equivalent to the tabular one when \( d = \|S\|\|A\| \) and the feature map is the indicator function for state-action pairs, for any \( d < \|S\|\|A\| \), a certain degree of transfer between state-action pairs becomes possible, with lower dimensionality leading to higher levels of transfer. 
While the feature map's dimensionality is evidently a crucial factor in MDP hardness, it is more challenging in this case to disentangle visitation complexity from estimation complexity, which is possible in the tabular setting. A reduction in the feature map's dimensionality has two effects: it decreases the number of state-action pairs the agent must visit (by increasing information transfer) and reduces the complexity of value function estimation (by requiring fewer parameters to be learned).
While the connection to linear regression theory offers theoretical advantages, this measure's practical application is severely limited in non-tabular settings as known feature maps that fit such strong assumptions are not available.

\textbf{Eluder dimension.}
Alternative approaches to quantifying the complexity of this setting can leverage the framework based on the eluder dimension introduced by \citet{osband2014model}.
The eluder dimension is generally defined as the longest possible sequence of tuples $(x_t, y_t)$ in a set such that it is not possible to estimate the function $x \mapsto y$ confidently.
Intuitively, this measure quantifies the ability of a functional class to fit an arbitrary sequence, the longer the sequence the harder it is to estimate the functional form, and it can be thought of as the reinforcement learning counterpart of the Vapnik–Chervonenkis dimension.
This measure has been further developed into the Bellman eluder dimension \citep{jin2021bellman} and the generalized rank \citep{li2022understanding}.
The eluder dimension could be particularly effective at capturing environment hardness because it both quantifies the visitation and estimation complexity.
However, this measure can be exponentially large even for a shallow neural network with ReLU activation \citep{dong2021provable}, meaning that the commonly used environments would all be assigned an infinite hardness value.

\textbf{Decision-estimation coefficient.}
A more general measure of hardness that can be applied to any sequential decision-making setting (e.g., bandits) is the decision-estimation coefficient \citep{foster2021instance}. Intuitively, the measure is the value of a game where the player must find a policy such that the regret is balanced with the estimation error for a worst-case problem instance. While the measure may be finite for environments of interest, it is not tractable to compute in practice.

\subsection{Open problems} \label{sec:open-problems}

\textbf{Intractability of non-tabular measures.}
The intractability of these hardness measures for relevant environments poses a key challenge for a practically useful theory of non-tabular hardness.
There are two potential directions forward.
The first is to develop measures that are less general and more instance-dependent.
Specific tailoring of them to structure assumptions of the environment of interests in the reinforcement learning community may allow the development of tractable measures.
The second one is to build practical approximations of the measures above.
For example, while an exact feature map of a linear MDP is not available, it may be possible to build neural surrogates similar to those for climatology \citep{kochkov2024neural} and physics \citep{eghbalian2023physics}.

\textbf{Representation learning.}
While not specifically addressed in the theory, the hardness of representation learning in non-tabular reinforcement learning is well renowned.
It is particularly clear to reinforcement learning researchers that an in-depth understanding of the representation complexity can lead to better-performing agents.
For example, \citet{dabney2021value, lyle2022understanding} shows that preventing feature collapse is very important for sparse-reward environments.
Also, \citet{nikishin2022primacy, d2022sample, schwarzer2023bigger} have shown that one key factor to improve the sample efficiency of reinforcement learning agents is to avoid the representation collapse of the Deep Q-Network (DQN) \citep{mnih2013playing} by resetting the network with random weights.
This is connected with feature collapse as explained by \citet{ni2024bridging}.
They show that several empirically validated techniques, e.g. stop-gradient and auxiliary loss functions, are effectively avoiding feature collapse, and so allow the policy of the agents to collapse to a sub-optimal local minimum.
A promising future direction can be the development of non-tabular measures of hardness specifically suited to capture the \textit{representation complexity}, which could complement the visitation and estimation complexity quantified by tabular measures.
However, it is important to note that different neural network architectures induce different inductive biases \citep{battaglia2018relational}.

\section{Benchmarking Reinforcement Learning} \label{sec:bench}

Section~\ref{sec:desiderata} proposes a set of desiderata that a principled benchmarking library for non-tabular reinforcement learning should possess.
Section \ref{sec:related} reviews the currently available benchmarking libraries highlighting gaps and limitations.
Finally, Section~\ref{sec:pharos} introduces \texttt{pharos}, a novel benchmarking library designed to address these shortcomings.

\subsection{Library desiderata} \label{sec:desiderata}

To systematically investigate non-tabular hardness and bridge the gap from theory to practice, a benchmarking library must satisfy several key requirements. These features are not merely convenient, but essential for conducting the kind of principled analysis this paper advocates for.

\textbf{Quantifiable Hardness.} A principled benchmark must be built on quantitative measures of difficulty. While the current state-of-the-art is limited to tabular metrics, the ability to compute them efficiently, even for environments with millions of states, is a foundational requirement. This enables a baseline for analysis and a means to test the very limits of these metrics.

\textbf{Integrated Agent Training.} To validate new algorithms or correlate hardness measures with agent performance, the library must seamlessly integrate with standard, state-of-the-art agent implementations. This ensures that findings are comparable and reproducible, and lowers the barrier for researchers to use the benchmark for their own evaluations.

\textbf{Broad Spectrum of Scale.} Effective research requires a spectrum of environments. The library should support the entire development pipeline, from rapid prototyping on small-scale, tabularly-solvable tasks to large-scale evaluations that challenge the limits of deep RL agents. This avoids the common problem of researchers needing to integrate multiple, disparate toolkits during a project.

\textbf{Parametric Environment Control.} Beyond providing a static set of environments, the library must allow for procedural generation over instance parameters (e.g., the size of a grid, the speed of cars). This capability is crucial for isolating specific hardness factors, e.g., systematically increasing a grid world's size to study visitation complexity while keeping the reward structure constant.

\textbf{Customizable Observations.} Finally, and most critically for non-tabular RL, the library must treat the observation function as a first-class, customizable component. To probe the impact of representation learning, an investigator must be able to train an agent on the exact same underlying MDP but with different observations—for example, a compact state vector versus a high-dimensional image. This ability to decouple the MDP's structural hardness from the representational challenge is essential for the analysis we present and is largely absent in existing benchmarks.

\subsection{Related works} \label{sec:related}

Several libraries have been developed to facilitate principled RL benchmarking, each with its own focus and limitations. Our work builds on their insights while addressing a critical gap in the study of non-tabular hardness.

\texttt{Bsuite} \citep{osband2020bsuite} provides a collection of small-scale experiments designed around expert intuitions of core RL challenges. While valuable for targeted analysis and interpretability, its scope is intentionally limited to simple tasks, and it does not allow for environment customization, precluding the systematic study of how factors like scale or representation impact hardness.

\texttt{Colosseum} \citep{conserva2022hardness} represents a step towards a benchmark based on computable tabular hardness metrics. This provides a theoretically grounded approach to hardness quantification. However, its focus remains on the tabular setting, and the computational cost of its analysis restricts its application to small-scale environments. It thus does not address the challenges of large-scale, non-tabular problems where representation learning is paramount.

More recently, \texttt{Bridge} \citep{laidlaw2023bridging} aims to connect tabular and non-tabular settings by providing simplified, low-dimensional representations of complex environments like Atari. While it provides access to interesting non-tabular dynamics and integrates with the popular Stable Baselines3 library, its implementation choices present challenges for customization, and it lacks the built-in analytical tools needed to compute hardness metrics.

As summarized in Table~\ref{tab:sample-table}, existing tools fall short of providing the unified framework needed for our investigation. \texttt{Bsuite} is based on intuition, \texttt{Colosseum} is grounded in theory but limited to the tabular world, and \texttt{Bridge} tackles non-tabular tasks but sacrifices customizability and deep analysis. \texttt{Pharos} is designed to fill this gap, uniquely combining scalable hardness computation with parametric control over both environment structure and agent observations.

\begin{table}[t]
\caption{Comparison of hardness-focused reinforcement learning benchmarking libraries. {\color{OliveGreen} Green}: Fully Supported, {\color{YellowOrange} Orange}: Partially Supported, {\color{BrickRed} Red}: Unavailable.}
\label{tab:sample-table}
\begin{center}
\begin{tabular}{lllll}
\multicolumn{1}{c}{Feature}
& \multicolumn{1}{c}{ \texttt{bsuite}}
& \multicolumn{1}{c}{ \texttt{colosseum}}
& \multicolumn{1}{c}{ \texttt{bridge}}
& \multicolumn{1}{c}{ \texttt{pharos} (Ours)}
\\ \hline \\
{Hardness measures} 
  & {\color{YellowOrange} Expert}
  & {\color{OliveGreen} Tabular} 
  & {\color{OliveGreen} Tabular} 
  & {\color{OliveGreen} Tabular} 
\\
{Agent training} 
  & {\color{YellowOrange} \texttt{dopamine}} 
  & {\color{YellowOrange} \texttt{acme}} 
  & {\color{OliveGreen} \texttt{SB3}} 
  & {\color{OliveGreen} \texttt{SB3}} 
\\
{Environments scale} 
  & {\color{YellowOrange} Small} 
  & {\color{YellowOrange} Small} 
  & {\color{OliveGreen} Small/Large} 
  & {\color{OliveGreen} Small/Large} 
\\
{Policy evaluation} 
  & {\color{BrickRed} Unavailable} 
  & {\color{BrickRed} Unavailable} 
  & {\color{YellowOrange} Partial} 
  & {\color{OliveGreen} Supported} 
\\
{Custom instances} 
  & {\color{BrickRed} Unavailable} 
  & {\color{OliveGreen} Supported} 
  & {\color{BrickRed} Unavailable} 
  & {\color{OliveGreen} Supported} 
\\
{Custom observations} 
  & {\color{BrickRed} Unavailable} 
  & {\color{OliveGreen} Supported} 
  & {\color{BrickRed} Unavailable} 
  & {\color{OliveGreen} Supported} 
\\
\end{tabular}
\end{center}
\end{table}

\subsection{Pharos} \label{sec:pharos}

\texttt{Pharos} is a novel benchmarking library that was specifically designed to satisfy the needs of researchers developing novel reinforcement learning agents and novel hardness measures.

\textbf{Capabilities.}
Designed for principled benchmarking, \texttt{pharos} enables the computation of tabular hardness measures and the associated value functions. It integrates Stable Baselines3 for agent training and natively supports common stochastic environment types like sticky actions \citep{machado2018revisiting} and action randomization \citep{conserva2022hardness}. Sticky actions introduce a non-zero probability of repeating the previous action, while action randomization resembles an environment-enforced epsilon-greedy exploration strategy. Computing value functions under sticky actions or action randomization are non-trivial, as the state value becomes dependent on the previous action. Further details on these challenges are provided in Appendix~\ref{app:scalability}.

\textbf{Scalability.}
\texttt{Pharos} constructs environments by building scalable tabular representations, easily supporting millions of states. While conceptually straightforward, building such large state spaces presents significant scalability challenges. The underlying algorithm performs a depth-first search, expanding the state space using the transition function from an initial state. The primary challenge lies in efficiently tracking visited states to avoid loops, as the state space can rapidly exceed memory capacity. For instance, representing states as 30-integer tuples (approximately 300 bytes each in Python) requires 30GB of memory for a million states. Further details and the algorithm are provided in Appendix~\ref{app:scalability} and Algorithm~\ref{alg:ssb}.

\begin{figure}[htbp]
  \centering
  \begin{subfigure}[b]{0.3\textwidth}
    \centering
    \includegraphics[width=0.9\textwidth]{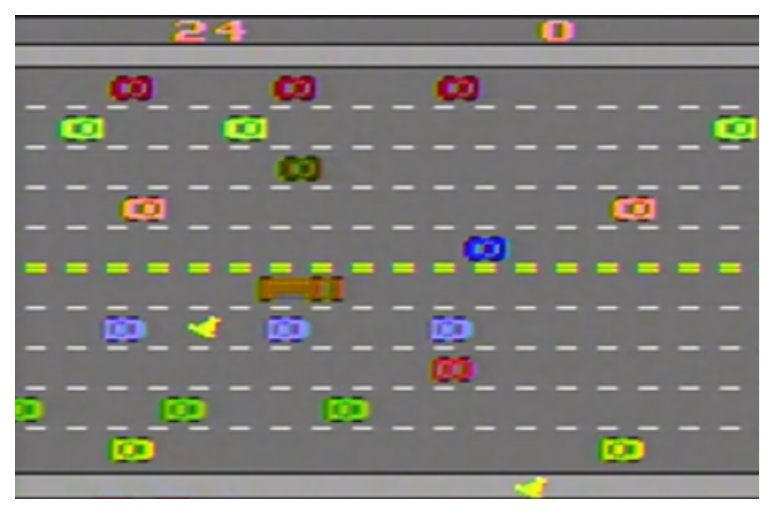}
    \caption{Original Atari.}
    \label{fig:freeway_original}
  \end{subfigure}
  \begin{subfigure}[b]{0.3\textwidth}
    \centering
    \includegraphics[width=0.9\textwidth]{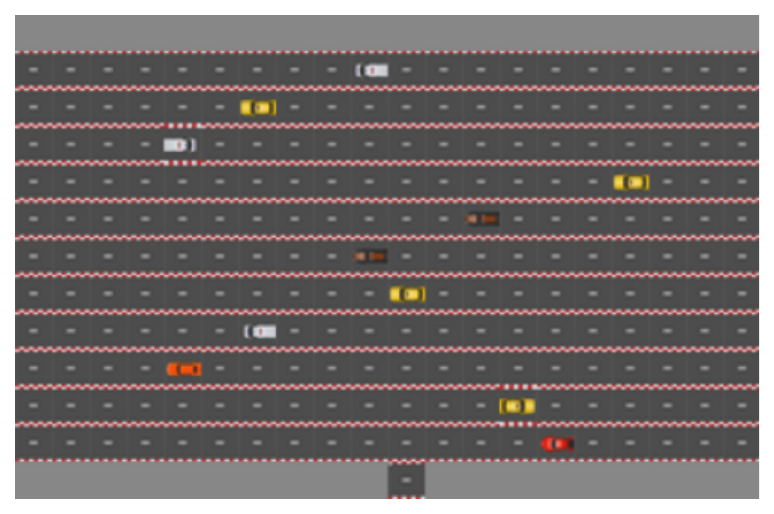}
    \caption{\texttt{Pharos} complex image. }
    \label{fig:freeway_complex}
  \end{subfigure}
  \begin{subfigure}[b]{0.3\textwidth}
    \centering
    \includegraphics[width=0.9\textwidth]{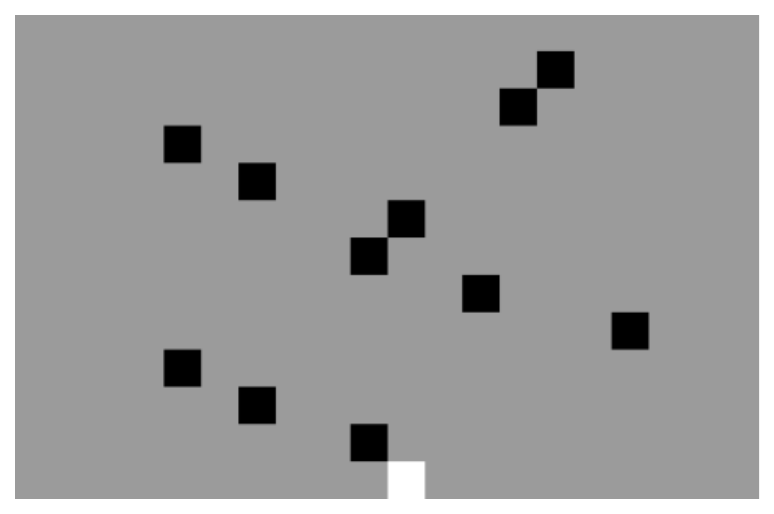}
    \caption{\texttt{Pharos} simple image.}
    \label{fig:freeway_simple}
  \end{subfigure}
  \caption{Atari like freeway available in \texttt{pharos} with different image representations.}
  \label{fig:freeway}
\end{figure}


\textbf{Available environments classes.}
The \texttt{simple\_grid} class of environments can be used to create grid world instances with different heights and widths.
It is also possible to add reward and penalty locations.
While the size of the class of environments grows unbounded with the height and width parameters, grid world environments are usually testbeds for small-scale testing of reinforcement learning agents.
Two specific grid world instances of different sizes are provided as \texttt{frozen\_lake} class of environments based on the widely used homonym environment.
While the scale of these environments is usually small, they can provide a significant challenge due to the penalty locations.
The \texttt{freeway} class of environments can be customized by modifying the speed at which the agent moves, the speed of the cars, the number of cars, and the length of lanes.
It is also possible, differently from the standard Atari game, to enforce that the agent restarts from the initial position at the bottom of the screen after hitting a car or to add a penalty for hitting a car.
This significantly increases the exploration challenge.
The \texttt{breakout} class of environments can be customized by changing the height, the number of columns, the number of rows with bricks, and the paddle size.
For each environment, we provide image representations and a vector representation which fully describes the state of the environment (similar to the RAM in Atari).
The image representation for freeway is shown in Figure~\ref{fig:freeway}, along with a comparison with the original Atari game representation.

\textbf{Limitations.}
A current limitation of \texttt{pharos} is the relatively small number of available environment classes.
While the existing classes offer some flexibility in customization, this limitation may restrict the diversity of the hardness analysis.
While \texttt{pharos} aims to simplify the process of creating additional environment, this still presents an obstacle to leveraging environments already available in other libraries.
Future development should prioritize expanding the variety of built-in environment classes and consider building bridges with other environment libraries to facilitate broader exploration and evaluation.

\section{Experiments} \label{sec:exp}

In Section \ref{sec:tab-vs-nontab} we examine the performance of the Q-learning algorithm in its tabular and non-tabular versions on small and large-scale environments to investigate how hard a simple and hard environment can be for a non-tabular and tabular agent, respectively.
In Section \ref{sec:visual}, we analyse the results of training DQN agents with varying representations to provide an example of the inductive biases of neural network.
In Section \ref{sec:corr} we perform a correlation study with several linear models to investigate whether or not tabular hardness measures can capture non-tabular hardness.

\textbf{Experimental setup.}
We conduct experiments across four environment classes: small-scale grid worlds and frozen lake variants (<100 states), and large-scale freeway and breakout variants (millions of states). For each class, we generate multiple instances by varying parameters (e.g., grid size, number of lanes) to create a diverse set of challenges. We train a DQN agent on each instance using both a simple image-based and a normalized vector-based representation, averaging results over five random seeds. Agents are trained for 50k steps in small-scale environments and 600k steps in large-scale ones, using hyperparameters optimized per environment class.

\textbf{Limitations.}
Our analysis focuses on a limited set of environments and primarily uses DQN to enable an in-depth qualitative study. While future work should broaden this scope to include more agents, environments, and representation types, we note that our core findings are not agent-specific. To validate this, we replicated our experiments with PPO and found a high performance correlation (0.81) with DQN. Full results for both agents are available in Appendix~\ref{app:agents_performances}.

\subsection{Tabular vs non-tabular performances} \label{sec:tab-vs-nontab}

\textbf{Motivation.}
As explained in Section~\ref{sec:hardness}, there are significant structural differences between the tabular and the non-tabular settings from a theoretical perspective.
In this section, we show how these differences play out in practice by comparing the performance of Q-learning in small and large-scale environments in its tabular version \citep{watkins1992q} and in its most widespread q-learning inspired agent DQN \citep{mnih2013playing}.

\begin{figure}[htbp]
  \centering
  \includegraphics[width=0.9\linewidth]{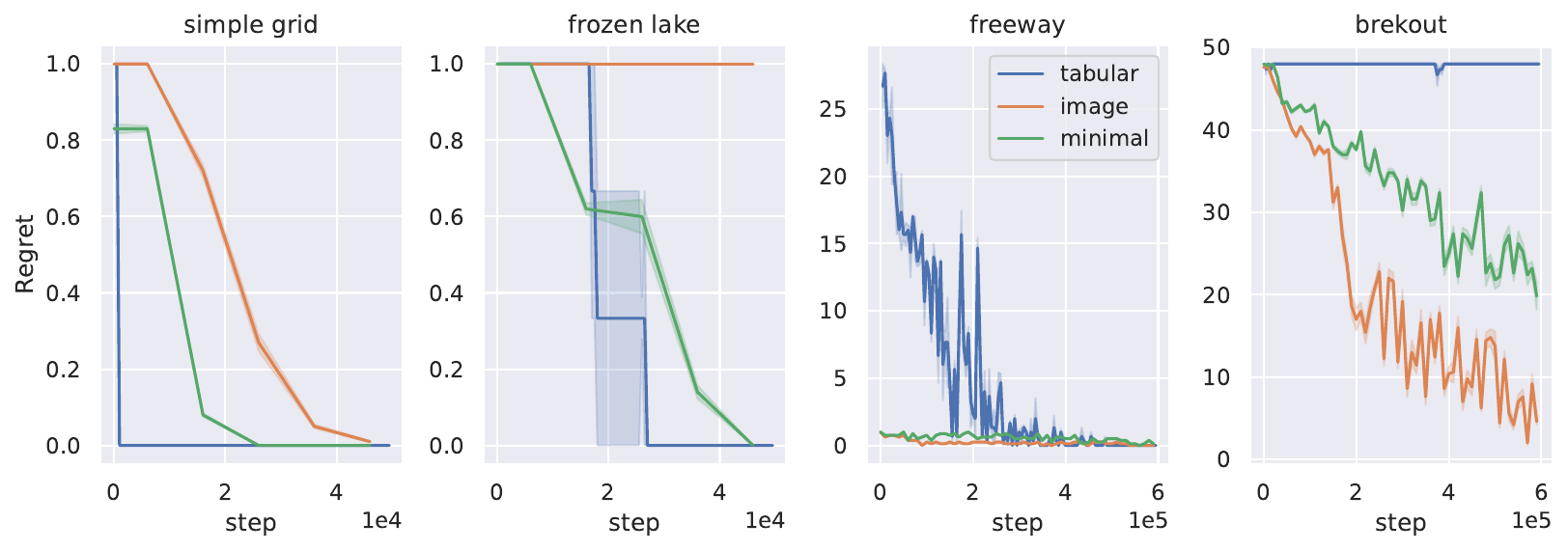}
  \caption{Regret plot for different environments available in pharos for the tabular case, for the case with a simple vector representation, and for the case of an image representation.}
  \label{fig:regret}
\end{figure}


\textbf{Results.}
Figures \ref{fig:regret} report the agent's performance in terms of cumulative regret, meaning that lower is better.
It is evident that in the smaller environments, the tabular agent performs significantly better than its non-tabular counterpart, where the DQN agent is not able to find the optimal policy for the frozen lake when provided with an image-based representation.
Instead, the non-tabular agent is better than the tabular counterpart for larger environments, with the tabular Q-learning not solving \texttt{breakout} in the given time limit.


\textbf{Discussion.}
While it is unsurprising that the tabular agent performs better in smaller environments than in larger ones and that its non-tabular counterpart struggles more with smaller environments, this distinction is crucial when discussing the applicability of tabular hardness measures in non-tabular settings.
The clear differences in what constitutes a difficulty for tabular versus non-tabular agents strongly suggest that these two settings are fundamentally distinct, making it unlikely that hardness measures from one can be reliably applied to the other.

\subsection{Representation learning challenge} \label{sec:visual}

\textbf{Motivation}
As mentioned in Section~\ref{sec:hardness}, different neural networks have different inductive biases.
To support this claim in practice, we compare the regret incurred by the DQN agent in the same environment instances when an image observation and a vector observation were provided.
The only difference in the agent is the feature extractor component, which is a convolutional neural network for the images and a fully connected neural network for vector observations.

\begin{figure}
  \centering
  \includegraphics[width=0.9\linewidth]{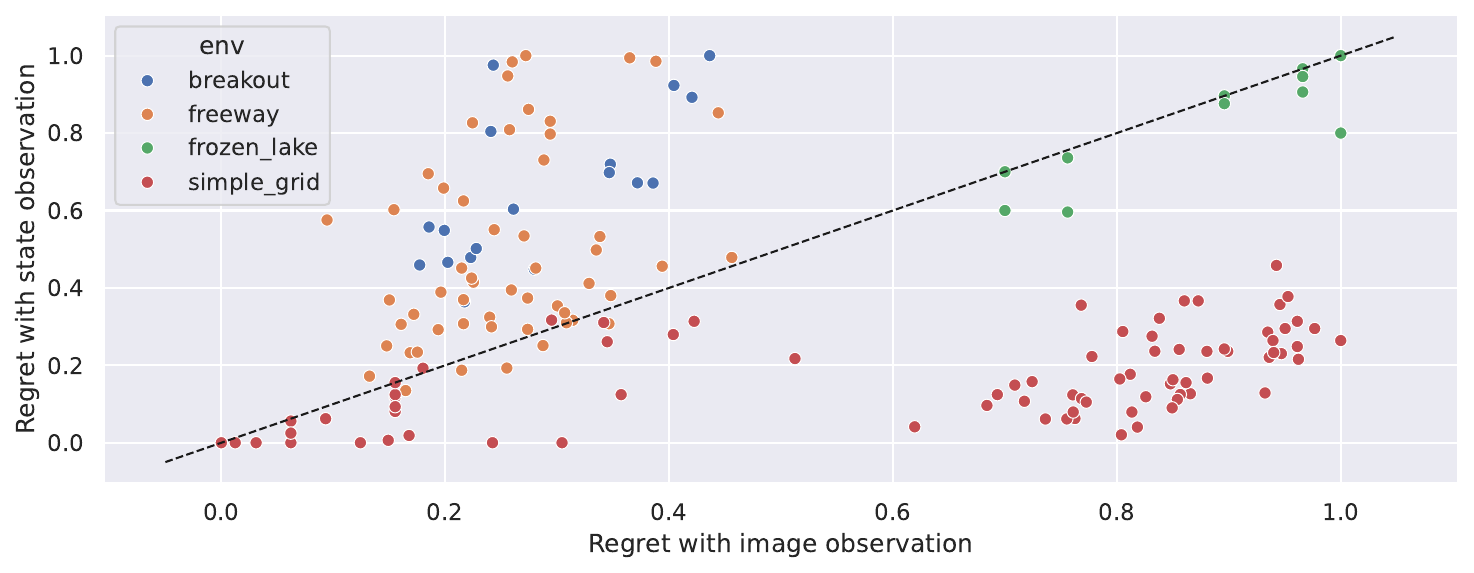}
  \caption{Normalized regret of DQN when provided with an image observation (x-axis) and with a state as observation (y-axis).
  DQN in breakout and freeway consistently incur more regret with a state observation compared with the image observation (points above the dotted line) whereas for a simple grid, the agent incurs more regret when provided with an image (points below the dotted line).
  For frozen lake, the regret is consistently high for both observation types.
  }
  \label{fig:representations}
\end{figure}

\textbf{Results.}
As shown in Figure~\ref{fig:representations}, the type of representation significantly changes the incurred regret.
However, we can notice that for the small-scale environments, $\texttt{frozen\_lake}$ and more evidently $\texttt{simple\_grid}$, the regret when the agent is presented with an image is higher compared to a vector (points below the dotted line).
The opposite instead holds for the large-scale environment \texttt{freeway} and \texttt{breakout} (points above the dotted line).

\textbf{Discussion.}
While there are no currently available tools to effectively pinpoint a potential explanation of the presented results.
This may be in line with the fact that neural networks, with their natural bias of producing smooth function, are not suited to be directly trained on tabular data \citep{beyazit2024inductive} such as the one that is the vector encoding the state of the environment.
While for \texttt{simple\_grid} and \texttt{freeway}, the state representation is a vector containing the x y coordinates, the state representation of \texttt{freeway} and \texttt{breakout} needs to contain many different objects resulting in a longer and tabular like representation.

\subsection{Correlation between hardness measures and non-tabular agents' performances} \label{sec:corr}

To better understand how well hardness measures capture the difficulty of environments, we fit linear models to predict the cumulative regret experienced by a DQN agent using these measures.
We choose to limit ourselves to linear models since they are highly interpretable. 

Although a single linear model applied across all environment instances (Section~\ref{sec:single}) does not yield an adequate fit, models tailored to specific representation types (Section~\ref{sec:representation-model}) and environment classes (Section~\ref{sec:class-models}) show improved performance.

\subsubsection{Single model} \label{sec:single}

The most general linear model is defined as,
\begin{equation} \label{eq:one_model}
\begin{aligned}
\mathrm{Regret}_{DQN} \sim\ & \alpha 
  + \beta_{1} \mathrm{representation}
  + \beta_{2} \mathrm{breakout}  
  + \beta_{3} \mathrm{freeway}
  + \beta_{4} \mathrm{frozen.lake}  \\
  &\ \ + \beta_{5} \log \mathrm{effective.horizon}
  + \beta_{6} \log \mathrm{sub.gaps}
  + \beta_{7} \log \mathrm{diameter},
\end{aligned}
\end{equation}
where the predictions are
$\mathrm{representation}$, which is a dummy variable encoding whether the observation type (image or vector) provided to the agent,
$\mathrm{breakout}$, $\mathrm{freeway}$, and $\mathrm{frozen.lake}$, which are dummy variables encoding the environment class \footnote{We omit explicitly adding a dummy variable for simple grid as it is implicitly captured in the constant $\alpha$.},
$\mathrm{effective.horizon}$, $\mathrm{sub.gaps}$, and $\mathrm{diameter}$, which are the tabular measure of hardness.

\begin{figure}[h!]
  \centering
  \includegraphics[width=0.9\linewidth]{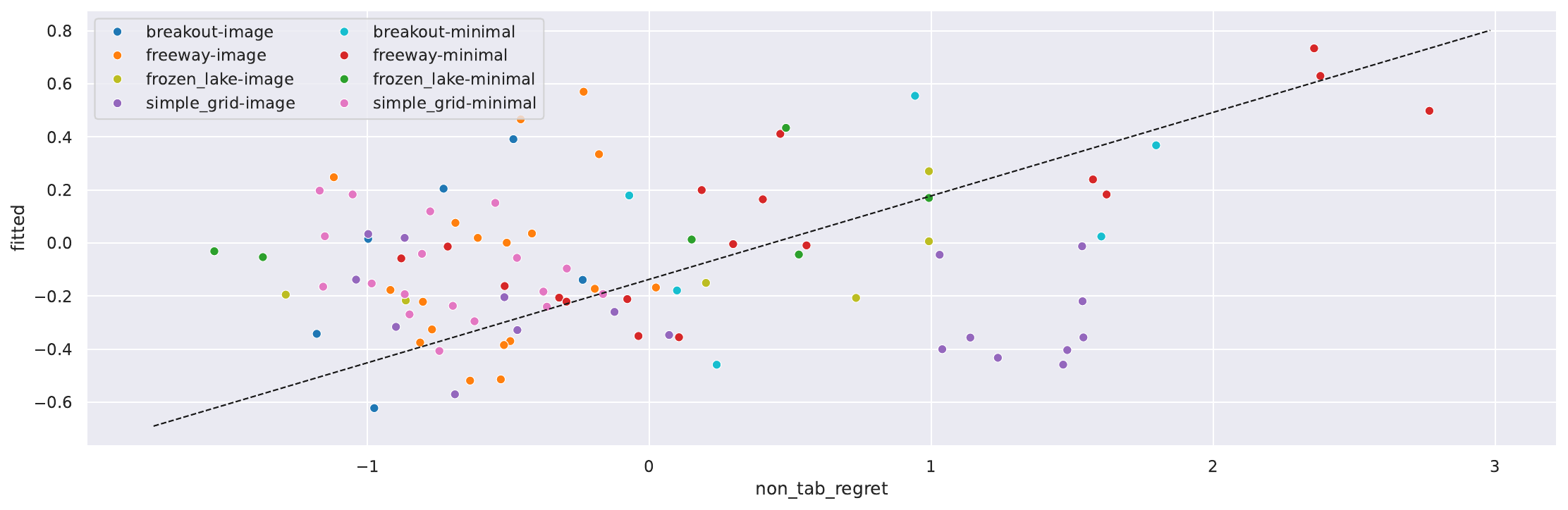}
  \caption{\textit{Fitted vs actual plot} for the single model (Equation \ref{eq:one_model}) where the non-tabular agent regret is inferred from tabular measure of hardness. The model is not able to fit well as shown by the fact that the points are not close to the identity diagonal line.}
  \label{fig:one_model}
\end{figure}

The $R^2$ score of this model is $0.09$, and the fitted vs actual plot is shown in Figure \ref{fig:one_model}.
The poor fit of this model and the absence of any statistical significance in the model coefficient indicate that tabular hardness measures are not able to capture the hardness of the non-tabular task in a way that generalizes across environment classes and representation types.

\subsubsection{Representation specific models} \label{sec:representation-model}

\begin{figure}[h!]
  \centering
  \includegraphics[width=0.7\linewidth]{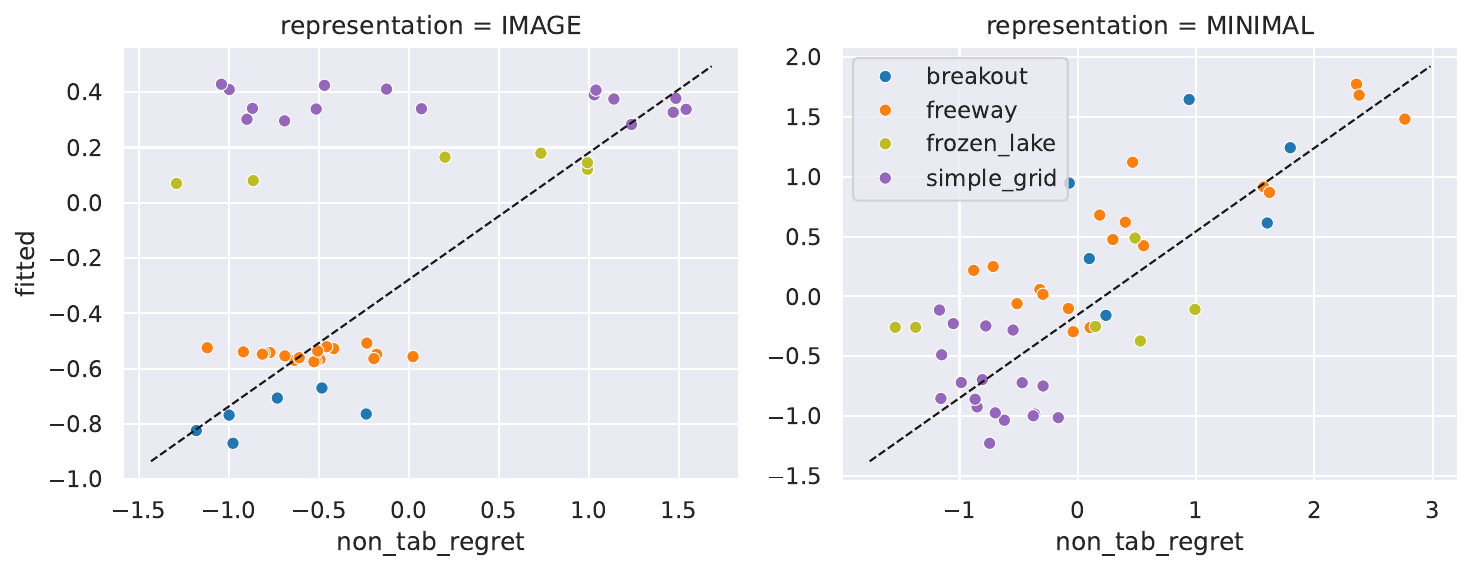}
  \caption{\textit{Fitted vs actual plot} for models separately fitted on the different representation types (Equation \ref{eq:rep_model}).
  While the fit for the vector representation is decent, the model is clearly not able to capture the case of image representations.}
  \label{fig:rep_model-label}
\end{figure}

Given that tabular measures are blind to observation complexity, we test whether their predictive power improves by fitting separate models for each representation type. The model is defined as:
\begin{equation} \label{eq:rep_model}
\begin{aligned}
\mathrm{Regret}_{DQN}^{(\mathrm{rep})} \sim\ & \alpha 
  + \beta_{2}^{(\mathrm{rep})} \mathrm{breakout}  
  + \beta_{3}^{(\mathrm{rep})} \mathrm{freeway}
  + \beta_{4}^{(\mathrm{rep})} \mathrm{frozen.lake}  \\
  &\ \ + \beta_{5}^{(\mathrm{rep})} \log \mathrm{effective.horizon}
  + \beta_{6}^{(\mathrm{rep})} \log \mathrm{sub.gaps}
  + \beta_{7}^{(\mathrm{rep})} \log \mathrm{diameter},
\end{aligned}
\end{equation}
The results reveal a stark contrast between the two settings (Figure~\ref{fig:rep_model-label}). For \textbf{image-based representations}, the model's fit was poor (R$^2$=0.3) with no statistically significant coefficients, indicating that tabular measures fail to explain agent performance. In contrast, for \textbf{vector-based representations}, the model was far more predictive (R$^2$=0.6), with most measures being significant. This divergence strongly suggests that the difficulty of vector-based tasks aligns more closely with traditional tabular hardness, whereas image-based tasks introduce a distinct representational challenge not captured by these metrics.

\subsubsection{Environment class specific models} \label{sec:class-models}

Finally, we investigate fitting linear models for each environment class to understand whether the hardness measures are more robust to the representation type in individual classes.
The models are in the form,
\begin{equation} \label{eq:env_model}
\begin{aligned}
\mathrm{Regret}_{DQN}^{(\mathrm{env})} \sim\ & \alpha 
  + \beta_{1}^{(\mathrm{rep})} \mathrm{representation} 
  + \beta_{2}^{(\mathrm{env})} \log \mathrm{effective.horizon} \\
  &\ \ + \beta_{3}^{(\mathrm{env})} \log \mathrm{sub.gaps}
  + \beta_{4}^{(\mathrm{env})} \log \mathrm{diameter},
\end{aligned}
\end{equation}
with a total number of parameters equal to $20$.

The results show that the predictive power and the importance of different factors vary dramatically across classes.
For {\texttt{frozen\_lake} (R$^2$=0.96)}, all tabular measures were significant predictors, while representation was not, indicating the model is robust in this near-tabular domain.
Conversely, for {\texttt{breakout} (R$^2$=0.92)}, both representation and suboptimality gaps were significant, demonstrating a joint influence of representational and estimation difficulty.
The model struggled with {\texttt{simple\_grid} (R$^2$=0.42)}, where the significant diameter metric failed to capture the crucial impact of reward penalties.
Most tellingly, for {\texttt{freeway} (R$^2$=0.67)}, only the representation coefficient was significant, suggesting that for this hard-exploration task, the challenge is almost entirely representational and not captured by the tabular metrics.
Overall, these environment-specific models (aggregate R$^2$=0.65) provided a much better fit than the single model (R$^2$=0.08) or the representation-specific models (R$^2$=0.48), but the inconsistent significance of predictors confirms that there is no universal recipe of tabular measures for non-tabular hardness.

\begin{figure}[h!]
  \centering
  \includegraphics[width=0.8\linewidth]{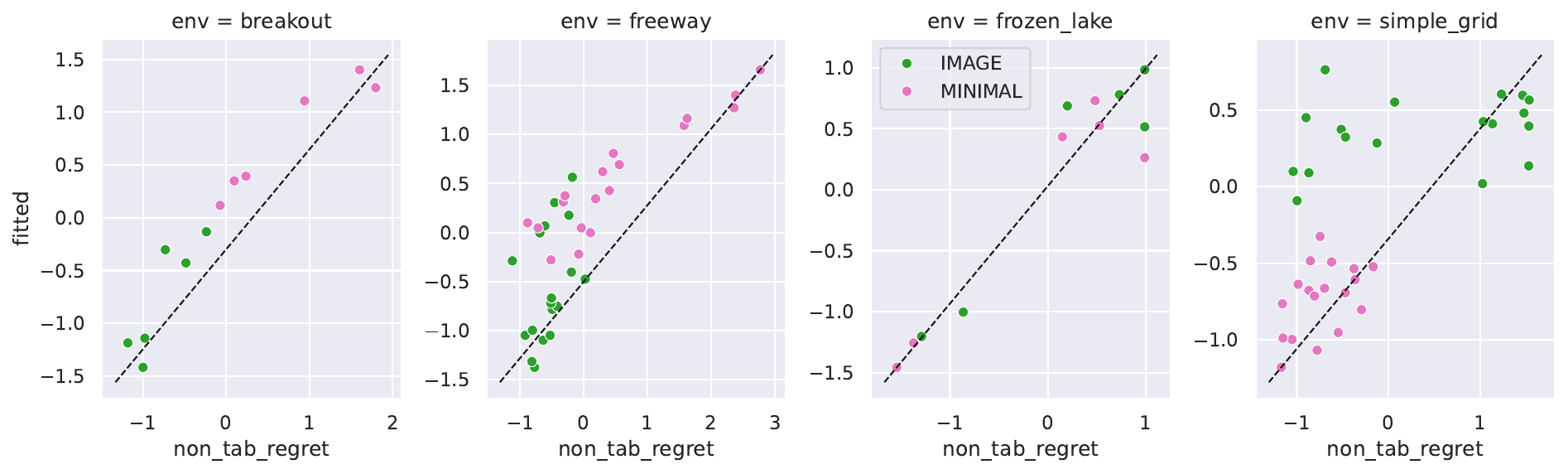}
  \caption{\textit{Fitted vs actual plot} for models separately fitted on the different environment classes. There is a relatively good fit between the classes with the exception of simple.grid.}
  \label{fig:env_model}
\end{figure}


\section{Conclusion} \label{sec:conclusion}

In this work, we challenged the implicit assumption that the theory of hardness from tabular RL can be directly applied to deep RL. Our investigation reveals a more complex reality: the difficulty of non-tabular environments is not merely a function of the underlying MDP structure, but is critically shaped by the agent's input representation. We find that representation hardness is often a more dominant factor than traditional measures of exploration or estimation complexity.

To conduct this analysis, we introduced \texttt{pharos}, a new benchmarking library designed for principled investigation into RL hardness. By enabling control over both environment parameters and observation types, \texttt{pharos} allowed us to systematically disentangle structural difficulty from representational difficulty. Our case study, leveraging DQN, demonstrates quantitatively that tabular hardness metrics are poor general predictors of deep RL agent performance. While they hold some predictive power in specific, controlled settings (e.g., within a single environment class with vector inputs), they fail to provide a universal measure of hardness.

This work signals a need for a paradigm shift in how we benchmark deep RL. Instead of relying on ad-hoc suites, we must move towards a more principled approach that explicitly accounts for representation. Our findings call for the development of new, tractable theories and metrics that capture representation complexity. \texttt{Pharos} provides an essential tool for this endeavor, offering a testbed to design and validate the next generation of hardness measures that will enable more meaningful and rigorous progress in reinforcement learning.

\clearpage
\bibliographystyle{unsrt} 
\bibliography{main}

\clearpage
\appendix

\section{Scalability} \label{app:scalability}

\paragraph{State space builder.}
The state space builder requires a transition function that computes the next state for any state and given action, the termination function that determines whether a state is terminal, and the starting state.
The \texttt{STATE} is a tuple of integers that fully describes the state of the environment.
For example, the position of the agent, the ball, and the brick in breakout.
The \texttt{ACTION} is an integer.
The algorithm starts by populating a queue of states to query with the starting state.
At every iteration, the first element is popped from the queue, and next states are computed for each action.
Any next state that has not been previously visited is then (i) added to the queue, (ii) its value is stored in a matrix on disk where the row corresponds to the state index and the columns to the state representation dimensionality, (iii) the index of the next state is stored in another on-disk matrix where the rows correspond to the previous state index and the columns to the action, and (iv) its value is stored in a hash map that keeps track of all the previously visited states.

\paragraph{Scalability challenges.}
The two key challenges in terms of memory are the queue, which can grow to a large extent when there are long sequences of novel states, and the hash map, which grows linearly with the number of states in the environment.
In practice, we have found the maximum size of the queue does not go over thousands of states, meaning that this occupies a relatively low proportion of the memory.
On the contrary, the increase of the size of the hash map is not avoidable, and, for large state space, it cannot be stored in memory.
To provide a reference for the potential size of this set we can assume that the state is represented by a tuple of $30$ integers, which occupies $300$ bytes of memory in Python.
For an environment with one hundred million states, this corresponds to more than $30$ GB of memory.
In order to scale to larger-than-memory state spaces, we therefore need an in-disk hash map that has high key retrieval speed and and high compression, with possibly slower IO speed.
While slower IO speed can be compensated with in-memory buffers, the retrieval speed is essential because the algorithm is constantly querying whether a next state has been seen before, and the compression rate is also necessary to avoid that the environment occupy too much disk space.
We have explored several options.
\texttt{sqlitedict} was immediately excluded due to slow retrieval speed.
\texttt{DiskDict} has a high key retrieval speed but does not natively support compression.
\texttt{ZODB} is reasonably fast in key retrieval and supports compression but \texttt{RocksDB} is better.
This library builds the hash map by storing sstables on disk.

\begin{algorithm}[tbh]
\caption{State Space Builder} \label{alg:ssb}
\KwIn{$starting\_state \gets \texttt{STATE}$ \tcp{Initial state of the environment}}
\KwIn{$transition\_function(s, action)$ \tcp{Function to compute the next state}}
\KwIn{$is\_terminal(s)$ \tcp{Function to check if a state is terminal}}

$queue \gets [starting\_state]$\; \tcp{Initialize queue with the starting state}
$visited\_states \gets \{\}$\; \tcp{Hash map to track visited states}
$state\_index \gets 0$\; \tcp{Index for the current state}

\While{$queue$ is not empty}{
  $current\_state \gets \texttt{pop}(queue)$\; \tcp{Pop first state from the queue}
  
  \If{$is\_terminal(current\_state)$}{
    \textbf{continue}\; \tcp{Skip if state is terminal}
  }
  
  \For{\textbf{each} $action$ \textbf{in} $ACTIONS$}{
    $next\_state \gets transition\_function(current\_state, action)$\; \tcp{Compute next state}
    
    \If{$next\_state$ \textbf{not in} $visited\_states$}{
      \texttt{append}(queue, $next\_state$)\; \tcp{Add new state to queue}
      \texttt{store\_state}(state\_index, $next\_state$)\; \tcp{Store state in disk matrix}
      \texttt{store\_transition}(state\_index, action, $next\_state$)\; \tcp{Store transition}
      $visited\_states[next\_state] \gets state\_index$\; \tcp{Mark state as visited}
      $state\_index \gets state\_index + 1$\; \tcp{Update state index}
    }
  }
}
\end{algorithm}

\paragraph{Reward function.}
After the state space has been built, the library goes on with the computation of the rewards.
While the rewards could be computed at the same time as building the state space, in practice, we have found it better to do it after because rewards can be computed in parallel based on the stored transitions, and we want to avoid further straining the intensive IO happening while building the state space.

\paragraph{Representation function.}
Similar to the rewards, the representation function of the environment can be computed after the state space has been fully created.
Given that the representation are only required during the interaction with the agent (no metric is computed based on them) we have found more beneficial to compute them on the fly to reduce the on-disk space requirements.
For an $64$ by $64$ image representation, the uncompressed space requirements for one hundred million states would be larger than $3$ TB.

\paragraph{Stochastic environments.}
Although incorporating support for state transition probability functions could be a future enhancement for the library, we believe that the substantial computational cost associated with this approach is not justified at present. 
Our assessment is based on the observation that most of the reinforcement learning community's work has focused on stochasticity types that can be derived from a deterministic transition function.
While the computation of environment properties under action randomization is largely the same as in non-randomized settings, sticky action introduces a notable change to the environment's structure. Specifically, in a sticky action environment, the next state depends not only on the previous state and the current action but also on the action taken in the previous step. This added dependency increases the complexity of computing environment properties, such as the optimal value function, which now requires a matrix with dimensions corresponding to the number of states by the number of actions, rather than a single vector of state lengths.
Given the previously described design, an important characteristic of the library is that the user can change the rewards and the representation without the necessity of building the full state space.
This is particularly important for researchers that are studying the representation learning challenges of deep reinforcement learning.

\clearpage
\section{Environment representations}

\clearpage
\section{Environment representations}

\begin{figure}[h!]
\centering
\begin{subfigure}[b]{0.23\textwidth}
\centering
\includegraphics[width=0.7\textwidth]{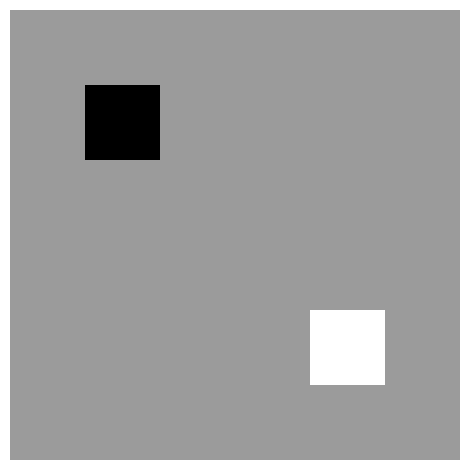}
\caption{(4x4)}
\label{fig:simple_grid_1}
\end{subfigure}
\hfill
\begin{subfigure}[b]{0.23\textwidth}
  \centering
  \includegraphics[width=0.7\textwidth]{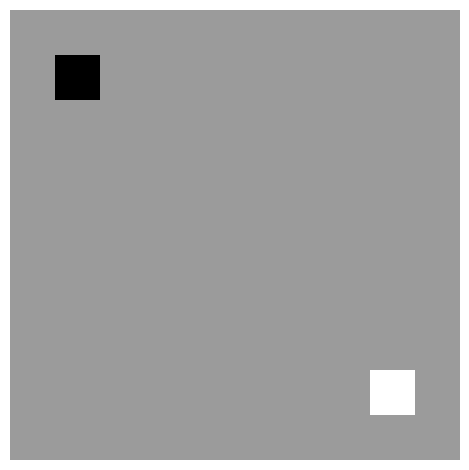}
  \caption{(8x8)}
   \label{fig:simple_grid_2}
  \end{subfigure}
  \begin{subfigure}[b]{0.23\textwidth}
    \centering
    \includegraphics[width=0.7\textwidth]{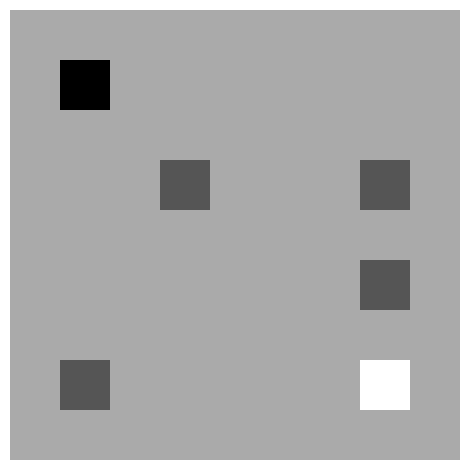}
    \caption{(4x4)}
    \label{fig:frozen_lake_1}
  \end{subfigure}
  \hfill
  \begin{subfigure}[b]{0.23\textwidth}
    \centering
    \includegraphics[width=0.7\textwidth]{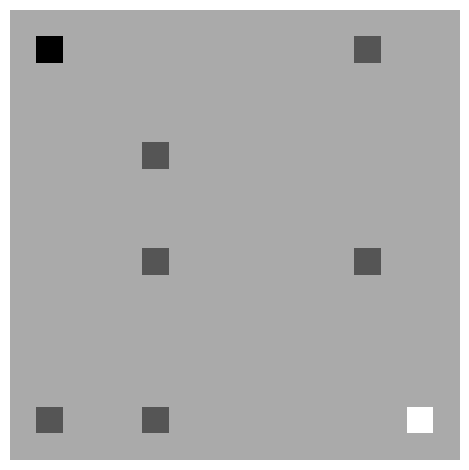}
    \caption{(8x8)}
    \label{fig:frozen_lake_2}
  \end{subfigure}
  
  \caption{Two different instances of SimpleGrid(height, width) and FrozenLake(height, width). The agent (black) has the simple task of reaching the goal (white). Frozen lake include holes (gray) that the agent needs to avoid.}
  \label{fig:frozen_lake_app}
\end{figure}

\begin{figure}[h!]
  \centering
  \begin{subfigure}[b]{0.19\textwidth}
    \centering
    \includegraphics[height=3cm]{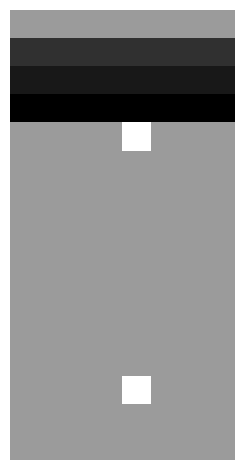}
    \caption{(14x8x1)}
    \label{fig:breakout_1}
  \end{subfigure}
  \hfill
  \begin{subfigure}[b]{0.19\textwidth}
    \centering
    \includegraphics[height=3cm]{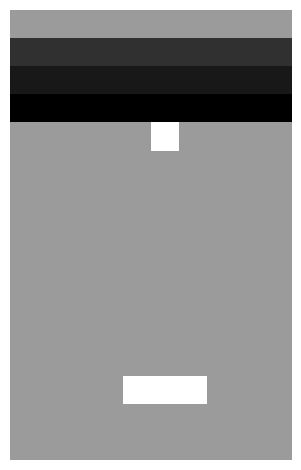}
    \caption{(14x10x3)}
    \label{breakout_2}
  \end{subfigure}
  \hfill
  \begin{subfigure}[b]{0.19\textwidth}
    \centering
    \includegraphics[height=3cm]{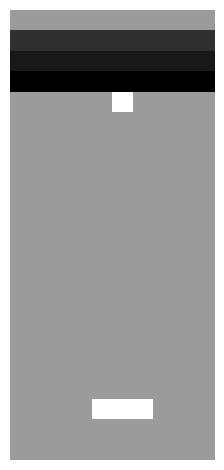}
    \caption{(20x10x1)}
    \label{breakout_3}
  \end{subfigure}
  \hfill
  \begin{subfigure}[b]{0.19\textwidth}
    \centering
    \includegraphics[height=3cm]{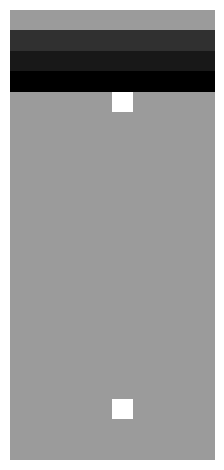}
    \caption{(20x10x3)}
    \label{fig:breakout_4}
  \end{subfigure}
  \hfill
  \begin{subfigure}[b]{0.19\textwidth}
    \centering
    \includegraphics[height=3cm]{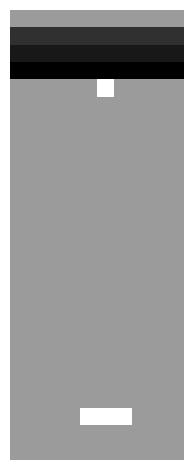}
    \caption{(24x10x3)}
    \label{fig:breakout_5}
  \end{subfigure}
  
  \caption{Five different instances of Breakout(height, width, paddle\_width). The black dots at the top of the figure represent the bricks that the agent needs to destroy by hitting the ball (white top dot) using the paddle (bottom dots).}
  \label{fig:breakout_app}
\end{figure}

\begin{figure}[h!]
  \centering
  \begin{subfigure}[b]{0.19\textwidth}
    \centering
    \includegraphics[height=3cm]{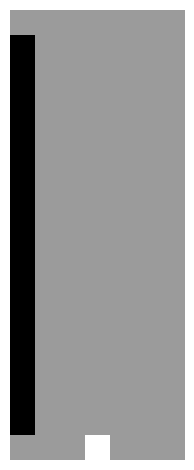}
    \caption{(16x7)}
    \label{fig:freeway_1}
  \end{subfigure}
  \hfill
  \begin{subfigure}[b]{0.19\textwidth}
    \centering
    \includegraphics[height=3cm]{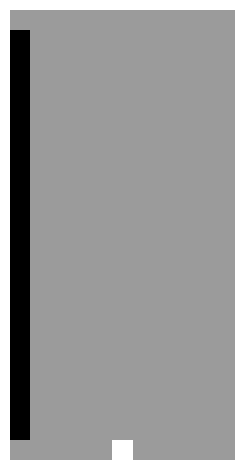}
    \caption{(20x11)}
    \label{freeway_2}
  \end{subfigure}
  \hfill
  \begin{subfigure}[b]{0.19\textwidth}
    \centering
    \includegraphics[height=3cm]{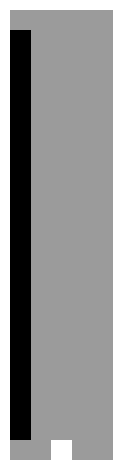}
    \caption{(20x5)}
    \label{fig:freeway_3}
  \end{subfigure}
  \hfill
  \begin{subfigure}[b]{0.19\textwidth}
    \centering
    \includegraphics[height=3cm]{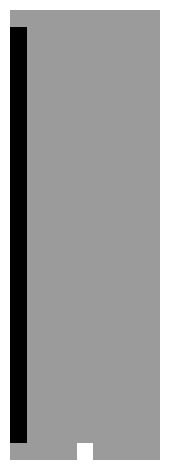}
    \caption{(25x9)}
    \label{fig:freeway_4}
  \end{subfigure}
  \hfill
  \begin{subfigure}[b]{0.19\textwidth}
    \centering
    \includegraphics[height=3cm]{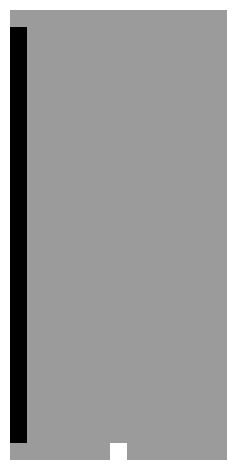}
    \caption{(25x13)}
    \label{fig:freeway_5}
  \end{subfigure}
  
  \caption{Five different instances of Freeway(num\_cars, lane\_length). The black dots on the left side of the figure represent cars whereas the white dot on the bottom is the agent. The goal of the agent is to reach the top by moving vertically without being hit by the cars moving horizontally.
  Users can modify the number of cars and the length of the lanes.
  Increasing the number of cars or reducing the lane length will make the environment more difficult, as the agent must navigate a higher number of moving vehicles within a more restricted space.}
  \label{fig:freeway_app}
\end{figure}

\clearpage
\section{Agent's performances} \label{app:agents_performances}

\begin{table}[h!]
\centering
\caption{Cumulative return of DQN and PPO agents on \texttt{simple grid} environments.}
\resizebox{0.85\textwidth}{!}{%
\begin{tabular}{llcccc}
\toprule
 & & \multicolumn{2}{c}{PPO} & \multicolumn{2}{c}{DQN} \\
\makecell[l]{environment\\parameter} & randomization & IMAGE & MINIMAL & IMAGE & MINIMAL \\
\midrule
\multirow[c]{3}{*}{{\small\makecell[l]{height=4, width=4}}} & None & $0.03 \pm 0.04$ & $0.78 \pm 0.20$ & $0.10 \pm 0.00$ & $1.00 \pm 0.00$ \\
 & random & $0.24 \pm 0.17$ & $0.89 \pm 0.00$ & $0.14 \pm 0.00$ & $0.92 \pm 0.00$ \\
 & stick & $0.00 \pm 0.01$ & $0.45 \pm 0.28$ & $0.22 \pm 0.00$ & $0.99 \pm 0.00$ \\
\cline{1-6}
\multirow[c]{3}{*}{{\small\makecell[l]{height=4,width=4}}} & None & $0.42 \pm 0.41$ & $0.86 \pm 0.19$ & $1.00 \pm 0.00$ & $1.00 \pm 0.00$ \\
 & random & $0.94 \pm 0.09$ & $1.00 \pm 0.00$ & $1.00 \pm 0.00$ & $1.00 \pm 0.00$ \\
 & stick & $0.10 \pm 0.07$ & $0.98 \pm 0.02$ & $1.00 \pm 0.00$ & $1.00 \pm 0.00$ \\
\cline{1-6}
\multirow[c]{3}{*}{{\small\makecell[l]{height=6,width=6}}} & None & $0.00 \pm 0.00$ & $0.57 \pm 0.41$ & $0.10 \pm 0.00$ & $0.99 \pm 0.00$ \\
 & random & $0.86 \pm 0.00$ & $0.91 \pm 0.03$ & $0.11 \pm 0.00$ & $0.98 \pm 0.00$ \\
 & stick & $0.03 \pm 0.03$ & $0.07 \pm 0.06$ & $0.17 \pm 0.00$ & $0.99 \pm 0.00$ \\
\cline{1-6}
\multirow[c]{3}{*}{{\small\makecell[l]{height=6,width=6}}} & None & $0.20 \pm 0.17$ & $0.97 \pm 0.01$ & $0.99 \pm 0.00$ & $1.00 \pm 0.00$ \\
 & random & $0.97 \pm 0.02$ & $0.99 \pm 0.01$ & $1.00 \pm 0.00$ & $1.00 \pm 0.00$ \\
 & stick & $0.15 \pm 0.14$ & $0.94 \pm 0.05$ & $0.99 \pm 0.00$ & $1.00 \pm 0.00$ \\
\cline{1-6}
\multirow[c]{3}{*}{{\small\makecell[l]{height=8,width=8}}} & None & $0.09 \pm 0.12$ & $0.44 \pm 0.36$ & $0.10 \pm 0.00$ & $0.99 \pm 0.00$ \\
 & random & $0.83 \pm 0.01$ & $0.89 \pm 0.03$ & $0.10 \pm 0.00$ & $0.98 \pm 0.01$ \\
 & stick & $0.00 \pm 0.00$ & $0.08 \pm 0.07$ & $0.12 \pm 0.00$ & $0.98 \pm 0.00$ \\
\cline{1-6}
\multirow[c]{3}{*}{{\small\makecell[l]{height=8,width=8}}} & None & $0.04 \pm 0.03$ & $0.66 \pm 0.38$ & $0.98 \pm 0.00$ & $0.99 \pm 0.00$ \\
 & random & $0.85 \pm 0.01$ & $1.00 \pm 0.00$ & $0.99 \pm 0.00$ & $1.00 \pm 0.00$ \\
 & stick & $0.00 \pm 0.00$ & $0.35 \pm 0.05$ & $0.99 \pm 0.00$ & $0.99 \pm 0.00$ \\
\bottomrule
\end{tabular}}
\label{tab:simple_grid_table}
\end{table}

\begin{table}[h!]
\centering
\caption{Cumulative return of DQN and PPO agents on \texttt{frozen lake} environments.}
\resizebox{0.85\textwidth}{!}{%
\begin{tabular}{llcccc}
\toprule
 & & \multicolumn{2}{c}{PPO} & \multicolumn{2}{c}{DQN} \\
\makecell[l]{environment\\parameter} & randomization & IMAGE & MINIMAL & IMAGE & MINIMAL \\
\midrule
\multirow[c]{3}{*}{{\small\makecell[l]{height=4,width=4}}} & None & $0.00 \pm 0.00$ & $0.11 \pm 0.16$ & $0.00 \pm 0.00$ & $0.98 \pm 0.01$ \\
 & random & $0.00 \pm 0.00$ & $0.64 \pm 0.03$ & $0.01 \pm 0.00$ & $0.71 \pm 0.01$ \\
 & stick & $0.00 \pm 0.00$ & $0.59 \pm 0.04$ & $0.00 \pm 0.00$ & $0.98 \pm 0.00$ \\
\cline{1-6}
\multirow[c]{3}{*}{{\small\makecell[l]{height=8, width=8}}} & None & $0.00 \pm 0.00$ & $0.06 \pm 0.06$ & $0.00 \pm 0.00$ & $0.96 \pm 0.00$ \\
 & random & $0.00 \pm 0.00$ & $0.72 \pm 0.09$ & $0.01 \pm 0.00$ & $0.89 \pm 0.01$ \\
 & stick & $0.00 \pm 0.00$ & $0.30 \pm 0.29$ & $0.00 \pm 0.00$ & $0.94 \pm 0.01$ \\
\bottomrule
\end{tabular}}
\label{tab:frozen_lake_table}
\end{table}

\begin{table}[h!]
\centering
\caption{Cumulative return of DQN and PPO agents on \texttt{freeway} environments.}
\resizebox{0.9\textwidth}{!}{%
\begin{tabular}{llcccc}
\toprule
 & & \multicolumn{2}{c}{PPO} & \multicolumn{2}{c}{DQN} \\
\makecell[l]{environment\\parameter} & randomization & IMAGE & MINIMAL & IMAGE & MINIMAL \\
\midrule
\multirow[c]{3}{*}{{\small\makecell[l]{lane length=11\\n cars=20\\player speed=1}}} & None & $-2.77 \pm 2.16$ & $0.01 \pm 0.78$ & $0.12 \pm 0.10$ & $-0.60 \pm 0.60$ \\
 & random & $-22.26 \pm 12.48$ & $-9.80 \pm 11.04$ & $-2.00 \pm 0.36$ & $-7.60 \pm 5.08$ \\
 & stick & $-0.06 \pm 0.05$ & $-0.08 \pm 0.17$ & $-0.11 \pm 0.05$ & $-0.44 \pm 0.35$ \\
\cline{1-6}
\multirow[c]{3}{*}{{\small\makecell[l]{lane length=11\\n cars=20\\player speed=2}}} & None & $0.00 \pm 0.00$ & $0.65 \pm 0.08$ & $-0.65 \pm 0.48$ & $-1.32 \pm 0.11$ \\
 & random & $-45.47 \pm 10.50$ & $-13.93 \pm 14.85$ & $-7.78 \pm 0.86$ & $-25.02 \pm 1.28$ \\
 & stick & $-3.01 \pm 4.25$ & $-1.89 \pm 1.67$ & $-2.97 \pm 0.57$ & $-11.38 \pm 0.36$ \\
\cline{1-6}
\multirow[c]{3}{*}{{\small\makecell[l]{lane length=13\\n cars=25\\player speed=1}}} & None & $0.20 \pm 0.59$ & $-0.24 \pm 0.91$ & $-0.11 \pm 0.35$ & $-0.62 \pm 0.42$ \\
 & random & $-12.30 \pm 11.27$ & $-4.68 \pm 4.69$ & $-1.60 \pm 0.54$ & $-2.63 \pm 0.74$ \\
 & stick & $-0.20 \pm 0.25$ & $0.24 \pm 0.37$ & $-1.14 \pm 0.18$ & $-0.37 \pm 0.30$ \\
\cline{1-6}
\multirow[c]{3}{*}{{\small\makecell[l]{lane length=13\\n cars=25\\player speed=2}}} & None & $-5.99 \pm 4.94$ & $-0.48 \pm 1.04$ & $-0.29 \pm 0.27$ & $-1.84 \pm 0.27$ \\
 & random & $-38.84 \pm 14.27$ & $-4.92 \pm 2.16$ & $-8.74 \pm 1.78$ & $-22.32 \pm 0.29$ \\
 & stick & $-5.34 \pm 5.15$ & $-0.22 \pm 0.31$ & $-2.89 \pm 0.65$ & $-10.69 \pm 3.65$ \\
\cline{1-6}
\multirow[c]{3}{*}{{\small\makecell[l]{lane length=5\\n cars=20\\player speed=1}}} & None & $-0.80 \pm 1.13$ & $-0.01 \pm 0.01$ & $-0.75 \pm 0.42$ & $-4.82 \pm 0.56$ \\
 & random & $-55.62 \pm 10.27$ & $-65.34 \pm 5.13$ & $-8.46 \pm 1.38$ & $-41.34 \pm 1.98$ \\
 & stick & $0.00 \pm 0.00$ & $-1.00 \pm 0.83$ & $-6.13 \pm 1.05$ & $-32.46 \pm 0.85$ \\
\cline{1-6}
\multirow[c]{3}{*}{{\small\makecell[l]{lane length=5\\n cars=20\\player speed=2}}} & None & $-7.27 \pm 5.17$ & $-0.01 \pm 0.02$ & $-2.42 \pm 0.65$ & $-7.76 \pm 0.22$ \\
 & random & $-122.50 \pm 29.65$ & $-106.40 \pm 4.32$ & $-56.41 \pm 1.56$ & $-83.72 \pm 2.16$ \\
 & stick & $-16.59 \pm 22.98$ & $-0.00 \pm 0.00$ & $-18.41 \pm 5.79$ & $-44.41 \pm 2.51$ \\
\cline{1-6}
\multirow[c]{3}{*}{{\small\makecell[l]{lane length=7\\n cars=16\\player speed=1}}} & None & $-5.80 \pm 4.81$ & $0.24 \pm 0.55$ & $0.33 \pm 0.14$ & $-2.27 \pm 1.36$ \\
 & random & $-2.15 \pm 1.58$ & $-25.99 \pm 17.00$ & $-3.09 \pm 0.89$ & $-1.40 \pm 0.27$ \\
 & stick & $-4.92 \pm 6.95$ & $0.18 \pm 0.25$ & $-0.14 \pm 0.27$ & $-0.06 \pm 0.08$ \\
\cline{1-6}
\multirow[c]{3}{*}{{\small\makecell[l]{lane length=7\\n cars=16\\player speed=2}}} & None & $-4.01 \pm 3.88$ & $0.41 \pm 0.30$ & $-0.18 \pm 0.28$ & $-4.23 \pm 0.40$ \\
 & random & $-37.76 \pm 7.44$ & $-11.99 \pm 12.33$ & $-8.72 \pm 1.40$ & $-6.45 \pm 1.20$ \\
 & stick & $0.00 \pm 0.00$ & $-0.51 \pm 0.73$ & $-3.41 \pm 1.43$ & $-10.96 \pm 5.76$ \\
\cline{1-6}
\multirow[c]{3}{*}{{\small\makecell[l]{lane length=9\\n cars=25\\player speed=1}}} & None & $-4.12 \pm 2.95$ & $-1.48 \pm 1.25$ & $-0.51 \pm 0.28$ & $-1.94 \pm 0.19$ \\
 & random & $-64.03 \pm 0.12$ & $-38.86 \pm 25.74$ & $-8.30 \pm 1.29$ & $-27.00 \pm 0.78$ \\
 & stick & $0.00 \pm 0.00$ & $-1.19 \pm 1.69$ & $-2.50 \pm 1.04$ & $-6.22 \pm 1.20$ \\
\cline{1-6}
\multirow[c]{3}{*}{{\small\makecell[l]{lane length=9\\n cars=25\\player speed=2}}} & None & $-2.46 \pm 1.75$ & $-0.65 \pm 0.54$ & $-1.14 \pm 0.24$ & $-2.04 \pm 0.55$ \\
 & random & $-76.75 \pm 20.22$ & $-38.05 \pm 37.94$ & $-20.54 \pm 5.08$ & $-34.51 \pm 4.82$ \\
 & stick & $-26.97 \pm 2.22$ & $-11.89 \pm 12.85$ & $-7.37 \pm 2.00$ & $-25.16 \pm 1.47$ \\
\cline{1-6}
\bottomrule
\end{tabular}}
\label{tab:freeway_table}
\end{table}

\begin{table}[h!]
\centering
\caption{Cumulative return of DQN and PPO agents on \texttt{breakout} environments.}
\begin{tabular}{llcccc}
\toprule
 & & \multicolumn{2}{c}{PPO} & \multicolumn{2}{c}{DQN} \\
\makecell[l]{Environment\\parameter} & Random & IMAGE & MINIMAL & IMAGE & MINIMAL \\
\midrule
\multirow[c]{3}{*}{{\small\makecell[l]{extra paddle width=0\\height=12,columns=10}}} & None & $11.89 \pm 8.61$ & $3.50 \pm 1.64$ & $49.84 \pm 0.56$ & $34.43 \pm 2.91$ \\
 & random & $5.27 \pm 0.21$ & $4.12 \pm 0.14$ & $11.17 \pm 0.43$ & $9.49 \pm 0.43$ \\
 & stick & $2.92 \pm 2.75$ & $6.82 \pm 0.27$ & $15.71 \pm 0.93$ & $12.72 \pm 0.41$ \\
\cline{1-6}
\multirow[c]{3}{*}{{\small\makecell[l]{extra paddle width=0\\height=12,columns=8}}} & None & $4.35 \pm 0.50$ & $9.04 \pm 4.02$ & $42.67 \pm 0.27$ & $35.09 \pm 1.17$ \\
 & random & $7.74 \pm 0.92$ & $2.17 \pm 0.03$ & $15.52 \pm 0.84$ & $13.73 \pm 0.53$ \\
 & stick & $5.67 \pm 2.37$ & $6.26 \pm 3.20$ & $21.67 \pm 1.42$ & $14.81 \pm 0.76$ \\
\cline{1-6}
\multirow[c]{3}{*}{{\small\makecell[l]{extra paddle width=0\\height=16,columns=8}}} & None & $2.02 \pm 0.02$ & $2.00 \pm 0.00$ & $36.90 \pm 0.38$ & $25.80 \pm 0.64$ \\
 & random & $5.73 \pm 0.26$ & $1.98 \pm 0.00$ & $11.80 \pm 0.10$ & $10.51 \pm 0.19$ \\
 & stick & $2.05 \pm 0.08$ & $2.00 \pm 0.00$ & $15.65 \pm 0.76$ & $13.43 \pm 0.81$ \\
\cline{1-6}
\multirow[c]{3}{*}{{\small\makecell[l]{extra paddle width=0\\height=20,columns=8}}} & None & $4.56 \pm 3.23$ & $7.99 \pm 6.70$ & $38.85 \pm 2.50$ & $19.15 \pm 2.33$ \\
 & random & $3.03 \pm 1.23$ & $2.68 \pm 1.73$ & $11.83 \pm 0.43$ & $8.68 \pm 0.19$ \\
 & stick & $1.47 \pm 2.08$ & $3.09 \pm 1.94$ & $17.15 \pm 0.36$ & $10.09 \pm 0.58$ \\
\cline{1-6}
\multirow[c]{3}{*}{{\small\makecell[l]{extra paddle width=0\\height=24,columns=8}}} & None & $2.67 \pm 1.72$ & $2.03 \pm 2.61$ & $35.00 \pm 0.69$ & $20.67 \pm 4.46$ \\
 & random & $3.78 \pm 0.43$ & $1.38 \pm 1.35$ & $10.89 \pm 0.72$ & $8.04 \pm 0.10$ \\
 & stick & $2.62 \pm 1.86$ & $2.66 \pm 0.57$ & $14.89 \pm 0.37$ & $8.92 \pm 0.40$ \\
\cline{1-6}
\multirow[c]{3}{*}{{\small\makecell[l]{extra paddle width=1\\height=12,columns=10}}} & None & $12.81 \pm 7.14$ & $14.46 \pm 8.27$ & $52.66 \pm 0.65$ & $38.64 \pm 1.93$ \\
 & random & $11.92 \pm 0.29$ & $7.24 \pm 4.06$ & $25.78 \pm 0.32$ & $26.41 \pm 0.36$ \\
 & stick & $12.94 \pm 1.91$ & $6.73 \pm 3.86$ & $29.25 \pm 1.71$ & $31.43 \pm 0.89$ \\
\cline{1-6}
\multirow[c]{3}{*}{{\small\makecell[l]{extra paddle width=1\\height=12,columns=8}}} & None & $24.36 \pm 3.72$ & $13.51 \pm 4.34$ & $43.85 \pm 1.14$ & $39.84 \pm 0.66$ \\
 & random & $16.13 \pm 0.27$ & $17.41 \pm 2.32$ & $28.83 \pm 0.86$ & $30.25 \pm 0.49$ \\
 & stick & $19.74 \pm 1.24$ & $16.48 \pm 0.97$ & $33.99 \pm 0.77$ & $34.42 \pm 0.92$ \\
\cline{1-6}
\multirow[c]{3}{*}{{\small\makecell[l]{extra paddle width=1\\height=16,columns=8}}} & None & $2.00 \pm 0.00$ & $2.00 \pm 0.00$ & $40.90 \pm 1.14$ & $33.70 \pm 2.46$ \\
 & random & $7.41 \pm 1.47$ & $5.06 \pm 4.36$ & $23.48 \pm 0.82$ & $21.59 \pm 0.82$ \\
 & stick & $3.57 \pm 2.23$ & $2.00 \pm 0.00$ & $30.50 \pm 0.57$ & $27.12 \pm 1.27$ \\
\cline{1-6}
\multirow[c]{3}{*}{{\small\makecell[l]{extra paddle width=1\\height=20,columns=8}}} & None & $8.34 \pm 4.36$ & $4.00 \pm 0.00$ & $42.31 \pm 0.34$ & $32.36 \pm 1.46$ \\
 & random & $8.98 \pm 0.58$ & $5.77 \pm 1.64$ & $24.54 \pm 0.76$ & $22.18 \pm 0.35$ \\
 & stick & $10.38 \pm 0.71$ & $4.76 \pm 1.07$ & $30.06 \pm 1.32$ & $28.05 \pm 1.52$ \\
\cline{1-6}
\multirow[c]{3}{*}{{\small\makecell[l]{extra paddle width=1\\height=24,columns=8}}} & None & $5.01 \pm 2.63$ & $14.54 \pm 11.14$ & $40.75 \pm 0.55$ & $31.28 \pm 1.77$ \\
 & random & $7.88 \pm 0.67$ & $7.03 \pm 3.19$ & $24.50 \pm 0.86$ & $23.44 \pm 0.78$ \\
 & stick & $2.27 \pm 0.39$ & $10.70 \pm 6.94$ & $28.81 \pm 0.72$ & $29.52 \pm 0.59$ \\
\bottomrule
\end{tabular}
\label{tab:breakout_table}
\end{table}

\end{document}